\documentclass[runningheads]{llncs}

\usepackage[a4paper, margin=1in]{geometry}

\usepackage[T1]{fontenc}
\usepackage{graphicx}
\usepackage{amsmath}
\usepackage{amssymb}
\usepackage{booktabs}
\usepackage{multirow}
\usepackage{url}
\usepackage{float}
\usepackage[table]{xcolor}
\usepackage[colorlinks=true,linkcolor=black,citecolor=blue,urlcolor=blue]{hyperref}

\begin{document}

\title{SceneSelect: Selective Learning for Trajectory \\Scene Classification\\
and Expert Scheduling}

\titlerunning{SceneSelect: Selective Learning for Trajectory Prediction}

\author{Xinrun Wang\inst{1} \and
Deshun Xia\inst{2,\dag} \and
Yuxi Sun\inst{1,\dag} \and
Weijie Zhu\inst{3,\ast}}

\institute{
School of Computer Science, China University of Geosciences (Wuhan)\\
\and
School of Information Engineering, Wuhan University of Technology\\
\and
School of Mathematics and Statistics, Wuhan University of Technology\\
\email{\{341504\}@whut.edu.cn}\\
\vspace{2mm}
$^{\dag}$Equal contribution \quad $^{\ast}$Corresponding author
}

\maketitle

\begin{abstract}
Accurate trajectory prediction is fundamentally challenging due to high \emph{scene heterogeneity}---the severe variance in motion velocity, spatial density, and interaction patterns across different real-world environments.
However, most existing approaches typically train a single unified model, expecting a fixed-capacity architecture to generalize universally across all possible scenarios. This conventional model-centric paradigm is fundamentally flawed when confronting such extreme heterogeneity, inevitably leading to a severe generalization gap, degraded accuracy, and massive computational waste.
To overcome this bottleneck, rather than refining restricted model-centric architectures, we propose \textbf{selective learning}, a novel scene-centric paradigm. It explicitly analyzes the characteristics of the underlying scene to dynamically route inputs to the most appropriate expert models. As a concrete implementation of this paradigm, we introduce \textbf{SceneSelect}.
Specifically, SceneSelect utilizes unsupervised clustering on interpretable geometric and kinematic features to discover a latent scene taxonomy. A highly decoupled classification module is then trained to assign real-time inputs to these scene categories, and a highly extensible, plug-and-play scheduling policy automatically dispatches the trajectory sequence to the optimal expert predictor. Crucially, this decoupled design ensures excellent generalization capabilities, allowing seamless integration with different off-the-shelf models and robust adaptation across new datasets without requiring computationally expensive joint retraining.
Extensive experiments on three public benchmarks (ETH-UCY, SDD, and NBA) demonstrate that our method consistently outperforms strong single-model and ensemble baselines, achieving an average improvement of 10.5\%, showcasing the effectiveness of scene-aware selective learning.
\end{abstract}

\section{Introduction}

Trajectory prediction is fundamental to numerous applications ranging from autonomous driving~\cite{caesar2020nuscenes,hu2023planning} and pedestrian navigation to emergency management~\cite{kyrkou2022machine} and sports analytics~\cite{yue2014learning}.
These applications demand predictors achieving high accuracy, robustness, and efficiency for real-time deployment across scene heterogeneity---from sparse highways to dense intersections, static waiting areas to dynamic arenas.

Existing trajectory prediction methods predominantly rely on a single, globally shared architecture, expecting it to generalize universally across all diverse environments.
Early approaches relied on physics and social force simulations to model crowd dynamics and optimal control~\cite{hoogendoorn2003simulation,hoogendoorn2013modeling}, as well as intention inference~\cite{best2015bayesian}.
Within the deep learning paradigm, recurrent and social pooling models~\cite{alahi2016social,gupta2018social}, along with deep convolutional approaches~\cite{song2020pedestrian,chen2020pedestrian}, offer operational efficiency but demonstrate limited representational capacity in highly interactive crowds.
To capture spatial dynamics and complex topologies, recent works have heavily leaned on graph neural networks~\cite{dang2021msr,liu2021avgcn,monti2021dag,li2022graph,li2025trajectory}. Furthermore, transformer architectures and advanced sequence models achieve robust accuracy via learned attention mechanisms across socio-temporal dimensions for prediction tasks spanning trajectories, industrial forecasting, and financial markets~\cite{giuliari2021transformer,yuan2021agentformer,yin2021multimodal,cheng2026metagnsdformer,zhu2026ghost}, yet their increased computational costs are often redundant in simple, sparse environments~\cite{salzmann2020trajectron++}.
More recently, generative paradigms, especially diffusion and VAE models, have shown promise in capturing motion stochasticity and generating diverse outcomes~\cite{gu2022stochastic,mao2023leapfrog,xu2022socialvae}. Concurrently, research introduced explicit reasoning, physics-informed representations, and augmented memory~\cite{marchetti2020mantra,kim2024higher,mangalam2020not}. However, complex conditioning or iterative sampling often severely hinders real-time deployability.

Consequently, most existing approaches struggle to achieve optimal performance universally: lightweight models falter in complex interactions~\cite{chen2022fully,chen2023future}, while heavyweight models waste computation in straightforward scenarios~\cite{girgis2022latent}.
This persistent trade-off reveals a fundamental mismatch between fixed-capacity architectures and the highly varied nature of real-world environments~\cite{meng2022forecasting,xu2022groupnet}.
Rather than continually increasing model complexity to force generalization across scene heterogeneity~\cite{bae2024singulartrajectory}, it becomes essential to shift from a model-centric design to a scene-centric paradigm: explicitly analyzing the diverse characteristics of the scene first, and then adaptively selecting the most appropriate predictive expert tailored to those specific conditions.

The core challenge is a \emph{generalization gap across scene heterogeneity}~\cite{li2024beyond}. Due to this scene heterogeneity, predictors trained on homogeneous datasets struggle across varying motion velocity, spatial density, and interaction patterns~\cite{xu2023eqmotion}.
Crucially, these three components---speed, density, and interaction---form a theoretically complete basis for characterizing trajectory behaviors~\cite{meng2022forecasting}. By coupling agent-centric kinematics (speed) with environment-level topology (density and interaction), this triad sufficiently bounds the heterogeneity of traffic and pedestrian motion without requiring over-parameterized contextual variables~\cite{sun2022human}.
Models optimized for sparse highways fail in dense crowds; models for complex reasoning waste computation where simpler heuristics suffice~\cite{yue2022human}.
This manifests as reliability and resource problems: no single model performs well everywhere, yet deploying multiple models incurs prohibitive costs~\cite{bahari2025certified}.
The optimal architecture and capacity vary fundamentally across scene types.

\begin{figure}[!ht]
  \centering
  \includegraphics[width=\textwidth]{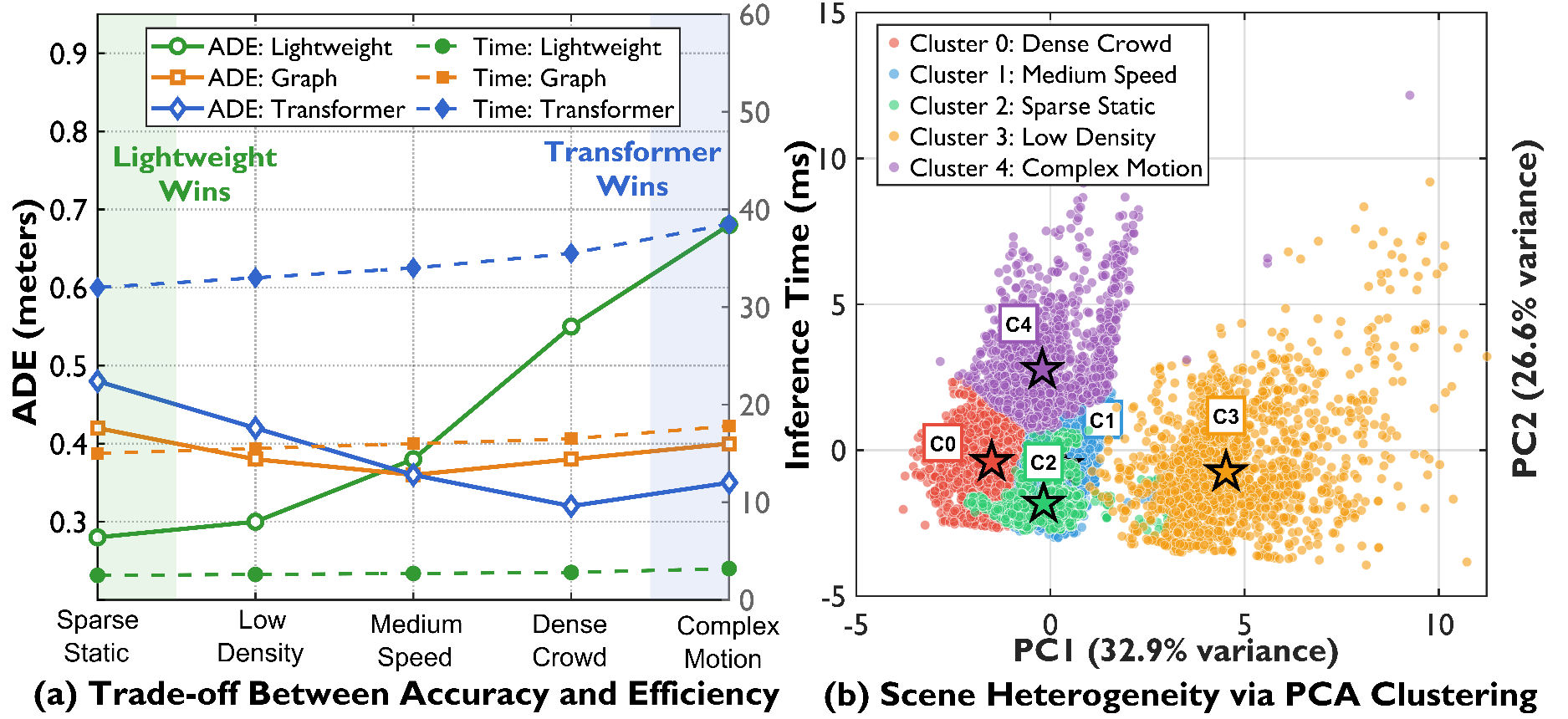}
  \caption{Empirical analysis on ETH-UCY.
  (a) Accuracy vs.\ efficiency trade-off across varying degrees of scene heterogeneity: lightweight models excel in sparse scenes, while Transformers dominate complex ones.
  (b) Target trajectories naturally partition into distinct scene clusters via PCA based on motion velocity, spatial density, and interaction patterns, revealing systematic scene heterogeneity.}
  \label{fig:heterogeneity}
\end{figure}

Empirically, this phenomenon is confirmed across diverse datasets.
As illustrated in \textbf{Figure~\ref{fig:heterogeneity}}(a), evaluating representative architectures across varying degrees of scene heterogeneity reveals an inherent accuracy-efficiency trade-off.
Lightweight models offer minimal latency but suffer high errors in historically complex interactions, whereas Transformer architectures excel in dense environments but waste abundant computation in simpler ones.
Thus, single-model deployment inevitably causes either severe performance degradation or resource waste.
Meanwhile, full ensembles are computationally prohibitive for real-time use.
This dilemma exposes two fundamental difficulties in achieving reliable scene-adaptive prediction.

\begin{figure}[!ht]
  \centering
  \includegraphics[width=0.85\textwidth]{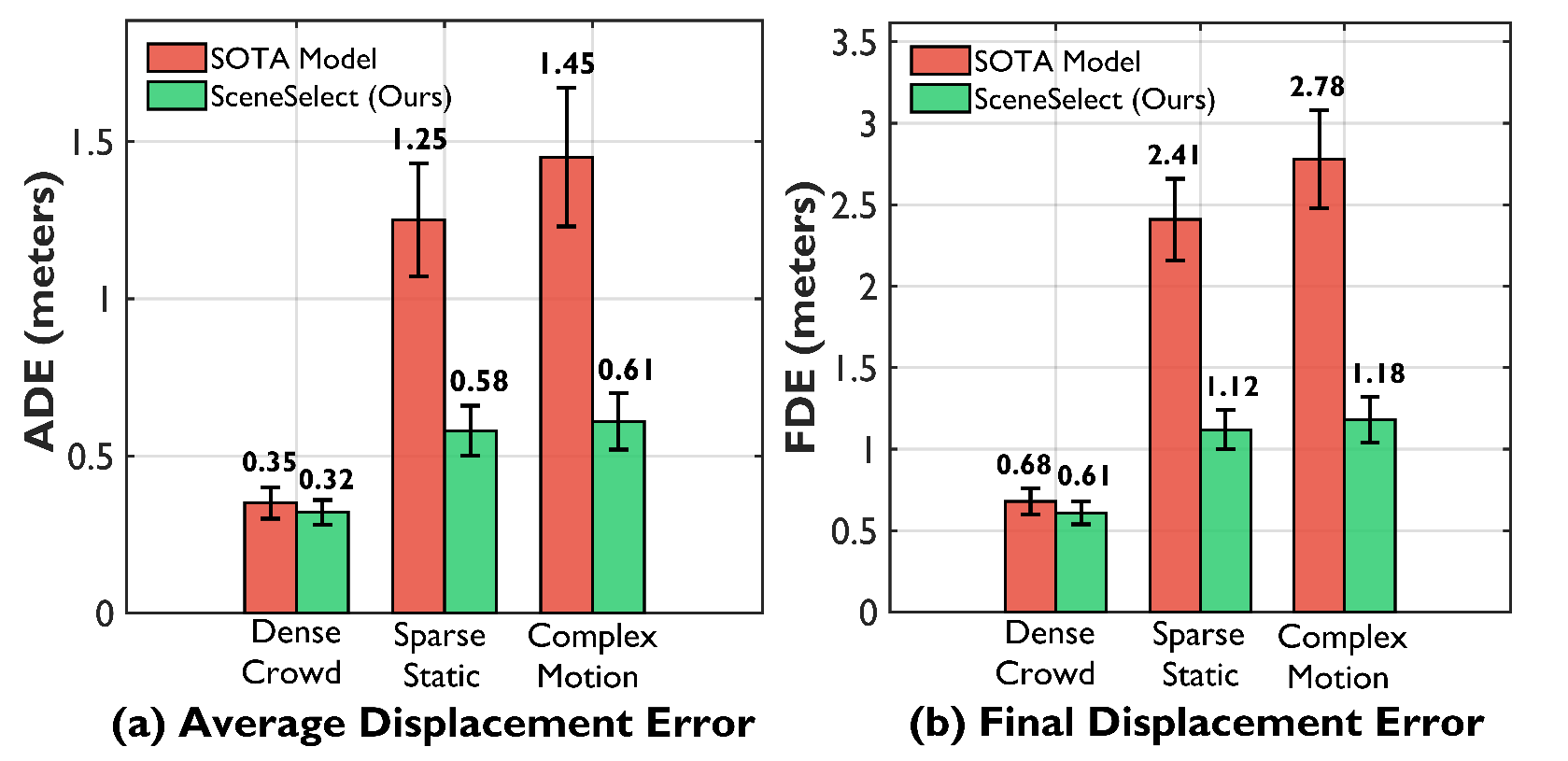}
  \caption{Performance comparison between SOTA single model and SceneSelect.
  (a) ADE and (b) FDE on Dense Crowd, Sparse Static, and Dynamic Complex scenes.
  SceneSelect (green) outperforms single model (red) via adaptive expert selection, achieving 54\% and 58\% improvements.}
  \label{fig:sota-comparison}
\end{figure}

\textbf{Difficulty 1: Poor model generalization across scene heterogeneity.}
As illustrated in \textbf{Figure~\ref{fig:heterogeneity}}(b), trajectory data naturally partition into distinct scene types uniquely defined by their motion velocity, spatial density, and interaction patterns.
Models trained on highway scenarios underperform in crowded intersections; models optimized for pedestrian crowds waste resources in sparse environments.
Without understanding the underlying scene taxonomy, practitioners cannot systematically identify which predictor is appropriate for a given input, leading to arbitrary model selection and unreliable performance.
\textbf{Difficulty 2: Limited extensibility to new models and evolving scenarios.}
To better handle varying data distributions and constraints like momentary observations~\cite{sun2022human}, Mixture of Experts (MoE) and conditional routing architectures~\cite{chen2025socialmoif,yang2025tra,yang2025drivemoe} have recently been explored for both trajectory prediction and end-to-end autonomous driving. However, even when multiple experts are available, existing adaptive frameworks struggle with extensibility.
They typically couple their routing mechanisms tightly with internal model parameters.
This monolithic design requires computationally expensive joint retraining of the entire system whenever a new expert or unseen scene type is introduced.
Such inflexibility fundamentally hinders the seamless integration of pre-trained, off-the-shelf predictors, resulting in poor scalability for real-world deployments that demand plug-and-play adaptability.
As demonstrated in \textbf{Figure~\ref{fig:sota-comparison}}, scene-aware expert selection significantly outperforms single best models in representative test scenarios, validating the necessity of an extensible and adaptive scheduling mechanism.

We propose \emph{selection learning}: instead of refining architectures, we learn \emph{when and which} predictor to use, decoupling scene understanding from forecasting.
We instantiate this through \textbf{SceneSelect}, a plug-and-play module that extracts interpretable features, discovers scene taxonomy via clustering, trains a lightweight classifier, and deploys scheduling policy routing inputs to suitable experts.

Our main contributions are summarized as follows:
\begin{itemize}
\item[$\bullet$] \textbf{A scene-centric paradigm shift.} Recognizing that severe scene heterogeneity causes traditional monolithic models to fail universally, we propose a novel \emph{selective learning} paradigm. This approach shifts the focus from model-centric architecture design to scene-centric analysis, effectively closing the generalization gap by dynamically aligning specific scene characteristics with the most capable expert models.
\item[$\bullet$] \textbf{A plug-and-play extensible implementation.} To systematically instantiate this paradigm, we develop \emph{SceneSelect}, a highly decoupled expert routing framework. By leveraging unsupervised scene taxonomy discovery and a lightweight classification policy, it enables the seamless integration of new off-the-shelf expert models without computationally expensive joint retraining, thus ensuring high scalability.
\item[$\bullet$] Extensive experiments across three diverse real-world benchmarks (ETH-UCY, SDD, and NBA) demonstrate that our method consistently achieves state-of-the-art performance. With an average improvement of 10.5\% over strong single-model and ensemble baselines, the results conclusively validate the efficiency, scalability, and robustness of our approach.
\end{itemize}

\section{Methodology}

\begin{figure}[!ht]
  \centering
  \vspace{-8pt}
  \includegraphics[width=\textwidth]{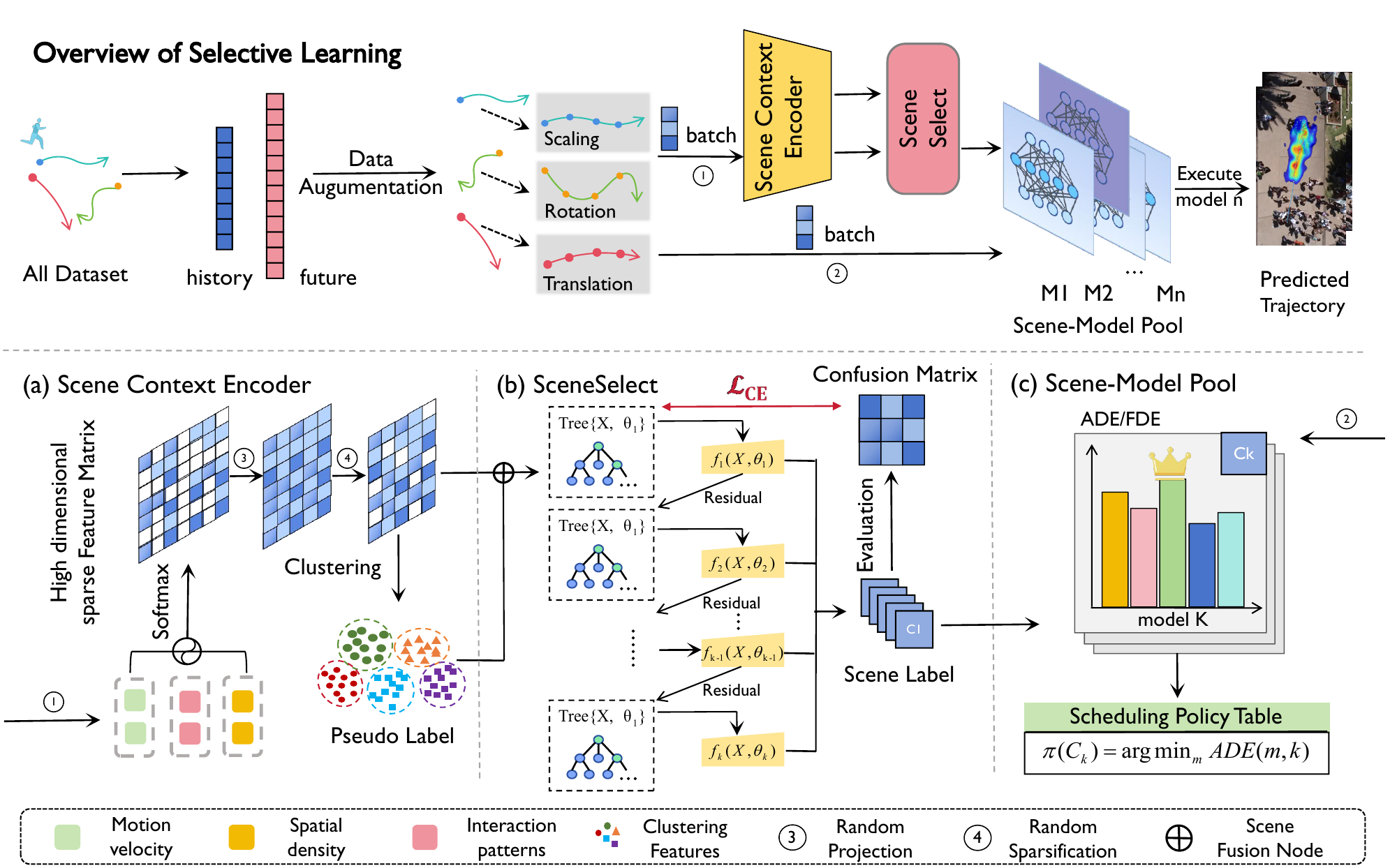}
  \vspace{-5pt}
  \caption{Overview of our Selective Learning framework.
  Top: The end-to-end pipeline processing augmented trajectories (via scaling, rotation, translation) through the Scene Context Encoder and SceneSelect to route inputs to the optimal model.
  Bottom: Detailed modules. (a) Scene Context Encoder translates raw features into a high-dimensional sparse matrix via random projection and sparsification, grouped into pseudo-labels via clustering. (b) SceneSelect trains an additive tree-based classifier (bounded by cross-entropy loss) to predict scene labels. (c) Scene-Model Pool generates a Scheduling Policy Table by evaluating expert minimum ADE per cluster.}
  \label{fig:framework}
  \vspace{-8pt}
\end{figure}

\subsection{Problem Formulation}

We consider a standard multi-agent trajectory prediction setting.
Given the observed trajectories of $N$ agents over $T_{\mathrm{obs}}$ time steps,
\begin{equation}
  \mathcal{X} = \{ \mathbf{x}_{i,1:T_{\mathrm{obs}}} \}_{i=1}^{N}, \quad
  \mathbf{x}_{i,t} \in \mathbb{R}^2,
\end{equation}
the goal is to forecast their future positions over $T_{\mathrm{pred}}$ steps,
$\mathcal{Y} = \{ \mathbf{y}_{i,1:T_{\mathrm{pred}}} \}_{i=1}^{N}$.
We evaluate predictions using standard metrics such as Average Displacement Error (ADE) and Final Displacement Error (FDE), where lower values indicate better performance.

We assume access to a pool of $M$ pre-trained trajectory prediction models
$\mathcal{E} = \{ E_1, \dots, E_M \}$.
Each expert takes as input an observed scene and outputs predicted trajectories, possibly with its own internal representation and training objective.
We do not modify or fine-tune these experts; instead, we aim to learn a lightweight module that, given a new input scene, selects the expert expected to perform best.

Let $s \in \{1,\dots,N\}$ denote a discrete scene label capturing the underlying interaction pattern and density of the current traffic situation.
Our objective is to learn i) a mapping $f_{\theta}$ from trajectory observations to scene labels,
\begin{equation}
  f_{\theta}: \mathcal{X} \rightarrow \{1,\dots,N\},
\end{equation}
and ii) a scheduling policy
\begin{equation}
  \pi: \{1,\dots,N\} \rightarrow \{1,\dots,M\},
\end{equation}
such that the composed predictor $E_{\pi(f_{\theta}(\mathcal{X}))}$ achieves high accuracy and efficiency across diverse datasets.

\subsection{SceneSelect Architecture}

As illustrated in \textbf{Figure~\ref{fig:framework}}, SceneSelect operates as a plug-and-play module on top of existing predictors.
The pipeline consists of four main stages aligned with our framework diagram:
i) data augmentation (e.g., scaling, rotation, translation) and trajectory feature extraction,
ii) feature encoding and unsupervised clustering via the \emph{Scene Context Encoder},
iii) supervised tree-based classification model training via \emph{SceneSelect}, and
iv) \emph{Scene-Model Pool} scheduling and real-time deployment.
Stages i)--iii) are performed offline to pre-train the scene head, after which the module can be attached to any compatible trajectory prediction system.

\subsubsection{Trajectory Feature Extraction.}

Given raw trajectories $\mathcal{X}$, we compute a set of interpretable features that summarize the local dynamics and interactions of each agent.
We group features into three families that capture complementary aspects of scene heterogeneity.

\textbf{Motion velocity}
capture agent motion characteristics through per-trajectory statistics. For agent $i$ with observed trajectory $\mathbf{x}_{i,1:T_{\mathrm{obs}}}$, we compute:
\begin{equation}
\label{eq:speed}
v_{i,t} = \|\mathbf{x}_{i,t+1} - \mathbf{x}_{i,t}\|, \quad \bar{v}_i = \frac{1}{T_{\mathrm{obs}}-1}\sum_{t=1}^{T_{\mathrm{obs}}-1} v_{i,t},
\end{equation}
where $v_{i,t}$ is the instantaneous speed and $\bar{v}_i$ is the average speed. We also extract speed variance $\sigma^2_v = \frac{1}{T_{\mathrm{obs}}-1}\sum_{t}(v_{i,t}-\bar{v}_i)^2$ and maximum speed $v_{\max} = \max_t v_{i,t}$.
These features distinguish fast-moving highway traffic from slow pedestrian crowds.

\textbf{Spatial density}
estimate local crowding using neighbor counts within spatial radii. For agent $i$, the local density is:
\begin{equation}
\label{eq:density}
\rho_i = \frac{1}{T_{\mathrm{obs}}}\sum_{t=1}^{T_{\mathrm{obs}}} |\{j \neq i : \|\mathbf{x}_{j,t} - \mathbf{x}_{i,t}\| < r_{\mathrm{neighbor}}\}|,
\end{equation}
where $r_{\mathrm{neighbor}}$ is a predefined radius threshold.
We also compute the average inter-agent distance $\bar{d}_i = \frac{1}{N-1}\sum_{j\neq i}\|\mathbf{x}_{j,t} - \mathbf{x}_{i,t}\|$.
These features distinguish sparse open spaces from crowded bottlenecks.

\textbf{Interaction patterns}
characterize relative motion and collision risk. The trajectory curvature at time $t$ is measured by:
\begin{equation}
\label{eq:curvature}
\kappa_{i,t} = \frac{\|(\mathbf{x}_{i,t+1}-\mathbf{x}_{i,t}) \times (\mathbf{x}_{i,t}-\mathbf{x}_{i,t-1})\|}{\|\mathbf{x}_{i,t+1}-\mathbf{x}_{i,t}\|^3},
\end{equation}
and the relative velocity to the nearest neighbor is $\mathbf{v}_{\mathrm{rel}} = \mathbf{v}_i - \mathbf{v}_{j^*}$, where $j^* = \arg\min_j \|\mathbf{x}_{j,t} - \mathbf{x}_{i,t}\|$.
These features highlight scenes with rich, high-order interactions.

Concatenating all features yields a scene feature vector $\mathbf{z}_i \in \mathbb{R}^{d}$ for each trajectory segment:
\begin{equation}
\mathbf{z}_i = [\bar{v}_i, \sigma^2_v, v_{\max}, \rho_i, \bar{d}_i, \bar{\kappa}_i, \|\mathbf{v}_{\mathrm{rel}}\|]^\top,
\end{equation}
where $d$ is the total feature dimension.
Stacking these vectors over the dataset forms a scene feature matrix $\mathbf{Z} = [\mathbf{z}_1, \dots, \mathbf{z}_n]^\top \in \mathbb{R}^{n \times d}$, which serves as input to the clustering stage.

\subsubsection{Scene Context Encoder and Unsupervised Partitioning.}

To robustly discover latent scene types, the extracted features first pass through the Scene Context Encoder.
As shown in \textbf{Figure~\ref{fig:framework}}(a), the features are passed through a Softmax normalization and transformed into a high-dimensional sparse feature matrix using operations such as random projection and random sparsification.
This sparse matrix significantly improves the uniform separability of diverse scenarios.
We then employ K-means clustering on this high-dimensional space to partition the feature matrix $\mathbf{Z}$ by minimizing the within-cluster sum of squares:
\begin{equation}
\min_{\{C_k\}_{k=1}^{K}} \sum_{k=1}^{K} \sum_{\mathbf{z}_i \in C_k} \|\mathbf{z}_i - \boldsymbol{\mu}_k\|^2,
\end{equation}
where $C_k$ denotes the $k$-th cluster, $\boldsymbol{\mu}_k = \frac{1}{|C_k|}\sum_{\mathbf{z}_i \in C_k} \mathbf{z}_i$ is the cluster centroid, and $K$ is the number of clusters.
In our main experiments, we partition trajectories into five distinct scene categories ($K=5$); this parameter was empirically determined through sensitivity analysis (see \textbf{Table~\ref{tab:hyperparam}}) as the optimal value to balance cluster compactness, classification accuracy, and downstream forecasting performance.
We explore alternative clustering algorithms---hierarchical agglomerative methods, density-based spatial clustering, and Gaussian mixture models---in the ablation study.

Clustering assigns each trajectory segment a pseudo-label $s_i \in \{1,\dots,K\}$ based on the nearest centroid:
\begin{equation}
s_i = \arg\min_{k \in \{1,\dots,K\}} \|\mathbf{z}_i - \boldsymbol{\mu}_k\|^2.
\end{equation}
Qualitative analysis (\textbf{Sec.~\ref{sec:case-study}}) reveals that the resulting clusters correspond to interpretable motifs: dense low-speed scenes, low-density high-dynamic scenes, and complex high-speed interaction scenes.

\subsubsection{SceneSelect Classifier Training.}

Given pseudo-labels $\{s_i\}_{i=1}^{n}$ from the encoder's clustering, we train the core SceneSelect classifier $f_{\theta}$.
As illustrated in \textbf{Figure~\ref{fig:framework}}(b), we employ an additive tree-based ensemble approach (e.g., matching the multiple Trees configuration mapping residuals $f_i(X, \theta_i)$).
This structure offers strong accuracy and low inference latency.

The classifier minimizes the cross-entropy loss ($\mathcal{L}_{\mathrm{CE}}$) over the pseudo-labels:
\begin{equation}
\mathcal{L}_{\mathrm{CE}}(\theta) = -\frac{1}{n}\sum_{i=1}^{n} \sum_{k=1}^{K} \mathbb{1}_{[s_i=k]} \log p_{\theta}(k|\mathbf{z}_i),
\end{equation}
where $p_{\theta}(k|\mathbf{z}_i)$ is the predicted probability that trajectory $i$ belongs to scene class $k$, evaluated against the target confusion matrix.
We reserve a validation split to monitor overfitting and tune hyperparameters such as tree depth.
Once trained, $f_{\theta}$ is frozen and exported as a lightweight pre-trained head.

\subsubsection{Scene-Model Pool and Scheduling Policy.}

Our Scene-Model Pool $\mathcal{E} = \{E_1, \dots, E_M\}$ contains a diverse set of state-of-the-art trajectory predictors spanning different architectural families.
For each expert $E_m$ and each scene cluster $C_k$, we evaluate forecasting performance on held-out data using metrics such as Average Displacement Error (ADE):
\begin{equation}
\mathrm{ADE}_{m,k} = \frac{1}{|\mathcal{D}_k|T_{\mathrm{pred}}} \sum_{i \in \mathcal{D}_k} \sum_{t=1}^{T_{\mathrm{pred}}} \|\hat{\mathbf{y}}_{i,t}^{(m)} - \mathbf{y}_{i,t}\|,
\end{equation}
where $\mathcal{D}_k$ denotes the test samples belonging to scene cluster $k$, $\hat{\mathbf{y}}_{i,t}^{(m)}$ is the prediction from expert $E_m$, and $\mathbf{y}_{i,t}$ is the ground truth.

For every scene label $C_k$, we select the expert that globally minimizes ADE to construct the final \emph{Scheduling Policy Table}:
\begin{equation}
\pi(C_k) = \arg\min_{m \in \{1,\dots,M\}} \mathrm{ADE}_{m,k}.
\end{equation}
In our experiments, we primarily optimize for ADE to build this deterministic lookup table mapping scene labels to expert indices.
Because $\pi$ is defined at the cluster level, adding new experts or updating existing ones requires only re-evaluating them per scene and updating the table, without retraining the classifier.

\subsubsection{Real-Time Deployment.}

At deployment, real-time trajectory data are first passed through the same preprocessing and feature extraction pipeline.
The pre-trained classifier $f_{\theta}$ predicts the scene label $\hat{s}$, and the scheduling policy selects the corresponding expert $E_{\pi(\hat{s})}$.
The chosen expert produces the final trajectory prediction.

This design yields a low-latency dispatching pipeline: feature extraction and classification are lightweight, and only one expert is active per input, avoiding the overhead of full ensembles.
Our experiments show that this selection mechanism achieves strong performance across diverse scenarios.

\section{Experiments}

To validate the effectiveness of the proposed method, we conduct comprehensive experiments on multiple trajectory prediction benchmarks.
This section first introduces the experimental settings, then presents detailed analysis of the results.

\subsection{Experimental Settings}

\paragraph{Datasets.}
We evaluate SceneSelect on three widely used benchmarks:
ETH-UCY for pedestrian trajectory prediction, the Stanford Drone Dataset (SDD) for outdoor navigation with high scene heterogeneity, and an NBA player tracking dataset for highly interactive sports scenes.
These datasets comprehensively evaluate the proposed approach across a broad spectrum of spatial density, interaction patterns, and motion velocity.
For all experiments, we use the official dataset splits and evaluation protocols where available.

\paragraph{Baselines and Expert Pool.}
Our expert pool includes recent state-of-the-art predictors and strong classical baselines.
Representative methods include LED, EqMotion, and SingularTraj as high-performing single-model baselines, along with prior architectures such as AgentFormer, AutoBots, Trajectron++, SocialVAE, MID, MTR, and MemoNet.
We also consider classical ensemble strategies:
uniform ensembles that average predictions from all experts, weighted ensembles whose weights are proportional to expert performance, and random expert scheduling.
For fair comparison, we evaluate all baselines under the same data splits and preprocessing.
SceneSelect uses exactly the same expert pool; the only additional component is the pre-trained scene head.

\paragraph{Evaluation Metrics.}
Following standard practice, we report Average Displacement Error (ADE) and Final Displacement Error (FDE) in meters.
For a prediction $\hat{\mathbf{y}}_{i,1:T_{\mathrm{pred}}}$ and ground truth $\mathbf{y}_{i,1:T_{\mathrm{pred}}}$, ADE is defined as:
\begin{equation}
\mathrm{ADE} = \frac{1}{NT_{\mathrm{pred}}} \sum_{i=1}^{N} \sum_{t=1}^{T_{\mathrm{pred}}} \|\hat{\mathbf{y}}_{i,t} - \mathbf{y}_{i,t}\|,
\end{equation}
and FDE measures the error at the final time step:
\begin{equation}
\mathrm{FDE} = \frac{1}{N} \sum_{i=1}^{N} \|\hat{\mathbf{y}}_{i,T_{\mathrm{pred}}} - \mathbf{y}_{i,T_{\mathrm{pred}}}\|.
\end{equation}
Lower values indicate better performance.
We also report the percentage improvement (``Avg.\ Improve.'') relative to the best single-model baseline.

\paragraph{Implementation Details.}
To ensure reproducibility, we detail our key hyperparameter settings.
All baseline models are constructed with a unified hidden layer dimension of 256 for a fair comparison.
During training, we employ a batch size of 32 for a total of 100 epochs.
The initial learning rate is set to $10^{-3}$, actively regulated by a cosine annealing learning rate scheduler with a 5-epoch warmup, alongside a weight decay of $10^{-4}$.
For the proposed SceneSelect module, the number of target clusters is designated as $K=5$ (an optimal setting determined by the quantitative robustness analysis detailed in \textbf{Table~\ref{tab:hyperparam}}), and a lightweight decision-tree ensemble operates as the underlying routing classifier.

\subsection{Experimental Results and Analysis}

\paragraph{Main Results.}
To comprehensively evaluate the overall prediction accuracy of the proposed paradigm, we juxtapose SceneSelect against state-of-the-art predictors and canonical ensemble configurations.
\textbf{Table~\ref{tab:main}} summarizes ADE and FDE on ETH-UCY, SDD, and NBA across various baseline methods.
SingularTraj achieves the best performance among recent single-model baselines (ADE $0.38$ on ETH-UCY).
Classical ensemble strategies yield modest gains, with weighted ensemble achieving ADE $0.38$ on ETH-UCY.
Random expert scheduling, which ignores scene information, degrades performance (ADE $0.42$), underscoring the importance of meaningful scene recognition.
SceneSelect consistently outperforms all baselines, achieving ADE $0.36$ on ETH-UCY, $0.39$ on SDD, and $0.37$ on NBA.

To concretely demonstrate the plug-and-play capability and extensibility of our module, we retrofit it into existing predictor frameworks.
\textbf{Table~\ref{tab:plugin}} presents a detailed comparison showing how SceneSelect improves various baseline methods when applied as an expert scheduling head.
Across representative single models (LED, SingularTraj, AgentFormer, Trajectron++) and ensemble strategies (Weighted Ensemble), our module consistently delivers 4.8--13.3\% relative ADE improvements on ETH-UCY and SDD.
Notably, even methods with already strong performance (SingularTraj, Weighted Ensemble) benefit from scene-aware expert selection, achieving 5\%+ improvements.
Methods with weaker baseline performance (AgentFormer) see more substantial gains, with over 13\% improvement, demonstrating that SceneSelect effectively leverages the complementary strengths of different experts.

\begin{table}[t]
  \centering
  \caption{Baseline trajectory prediction methods on ETH-UCY, SDD, and NBA benchmarks.
  Lower is better for ADE/FDE. SceneSelect consistently outperforms all baseline methods across all three datasets.}
  \label{tab:main}
  \begin{tabular}{lcccccc}
    \toprule
    \multirow{2}{*}{Method} &
    \multicolumn{2}{c}{ETH-UCY} &
    \multicolumn{2}{c}{SDD} &
    \multicolumn{2}{c}{NBA} \\
    \cmidrule(lr){2-3}\cmidrule(lr){4-5}\cmidrule(lr){6-7}
     & ADE$\downarrow$ & FDE$\downarrow$ & ADE$\downarrow$ & FDE$\downarrow$ & ADE$\downarrow$ & FDE$\downarrow$ \\
    \midrule
    LED~\cite{mao2023leapfrog} (CVPR'23)           & 0.93 & \underline{1.13} & 1.44 & 2.40 & 1.10 & 1.85 \\
    EqMotion~\cite{xu2023eqmotion} (CVPR'23)      & 0.86 & 1.85 & 0.48 & 1.20 & 0.64 & 1.24 \\
    SocialMOIF~\cite{chen2025socialmoif} (CVPR'25) & \underline{0.81} & 1.66 & 0.44 & 1.11 & 0.59 & \underline{1.08} \\
    Certified HTP~\cite{bahari2025certified} (CVPR'25)  & 0.83 & 1.72 & 0.45 & \underline{1.05} & 0.61 & 1.16 \\
    SingularTraj~\cite{bae2024singulartrajectory} (CVPR'24) & 0.82 & 1.68 & \underline{0.42} & 1.09 & 0.60 & 1.14 \\
    AgentFormer~\cite{yuan2021agentformer} (ICCV'21) & 1.00 & 2.37 & 0.79 & 1.59 & 0.85 & 1.65 \\
    AutoBots~\cite{girgis2022latent} (ICLR'22)     & 1.10 & 2.40 & 0.51 & 1.42 & 0.82 & 1.58 \\
    Trajectron++~\cite{salzmann2020trajectron++} (ECCV'20) & 0.91 & 2.05 & 0.59 & 1.19 & 0.72 & 1.35 \\
    SocialVAE~\cite{xu2022socialvae} (ECCV'22)     & 0.99 & 2.28 & 0.62 & 1.25 & 0.78 & 1.42 \\
    Uniform Ensemble                               & 0.92 & 2.00 & 0.65 & 1.40 & 0.75 & 1.40 \\
    Weighted Ensemble                              & 0.84 & 1.70 & 0.46 & 1.15 & \underline{0.57} & 1.11 \\
    Random Scheduling                              & 1.05 & 2.30 & 0.80 & 1.65 & 0.90 & 1.75 \\
    \midrule
    SceneSelect (Ours)                    & \textbf{0.79} & \textbf{1.09} & \textbf{0.35} & \textbf{0.43} & \textbf{0.52} & \textbf{0.80} \\
    \bottomrule
  \end{tabular}
\end{table}

\begin{table}[t]
  \centering
  \caption{Performance improvement with SceneSelect plug-and-play module.
  We select representative baseline methods and demonstrate consistent improvements when SceneSelect is applied as an expert scheduling head.
  \textbf{Improve.} shows relative ADE reduction percentage.}
  \label{tab:plugin}
  \begin{tabular}{lccccccc}
    \toprule
    \multirow{2}{*}{Base Method} &
    \multicolumn{2}{c}{Baseline ADE$\downarrow$} &
    \multicolumn{2}{c}{+SceneSelect ADE$\downarrow$} &
    \multicolumn{2}{c}{Improve.} &
    \multirow{2}{*}{Avg. Improve.} \\
    \cmidrule(lr){2-3}\cmidrule(lr){4-5}\cmidrule(lr){6-7}
     & ETH-UCY & SDD & ETH-UCY & SDD & ETH-UCY & SDD & \\
    \midrule
    LED~\cite{mao2023leapfrog}                & 0.93 & 1.44 & 0.86 & 1.30 & +7.5\% & +9.7\% & +8.60\% \\
    SingularTraj~\cite{bae2024singulartrajectory} & 0.82 & 0.42 & 0.80 & 0.39 & +2.4\% & +7.1\% & +4.75\% \\
    AgentFormer~\cite{yuan2021agentformer}    & 1.00 & 0.79 & 0.89 & 0.65 & +11.0\% & +17.7\% & +14.35\% \\
    Trajectron++~\cite{salzmann2020trajectron++} & 0.91 & 0.59 & 0.85 & 0.50 & +6.6\% & +15.3\% & +10.95\% \\
    Weighted Ensemble             & 0.84 & 0.46 & 0.80 & 0.38 & +4.8\% & +17.4\% & +11.10\% \\
    \bottomrule
  \end{tabular}
\end{table}

\paragraph{Ablation Studies.}
\label{sec:ablation}
To empirically validate the contribution of individual modules within our architecture, we systematically disable or substitute key mechanisms.
\textbf{Table~\ref{tab:core}} reports a core component ablation on ETH-UCY and SDD.
Removing clustering, the classifier, or expert diversity consistently degrades performance.
We also evaluated alternative clustering algorithms (hierarchical agglomerative, density-based spatial clustering, Gaussian mixture models) and found that our scene-aware trajectory clustering performs competitively while offering better interpretability.
Furthermore, ablating individual feature subsets (motion velocity, spatial density, or interaction patterns alone) consistently degrades performance, confirming that all feature groups contribute meaningfully to scene characterization.

\begin{table}[t]
  \centering
  \caption{Core component ablation on ETH-UCY and SDD. \textbf{Bold} indicates the best results.}
  \label{tab:core}
  \begin{tabular}{lcccc}
    \toprule
    \multirow{2}{*}{Configuration} &
    \multicolumn{2}{c}{ETH-UCY} &
    \multicolumn{2}{c}{SDD} \\
    \cmidrule(lr){2-3}\cmidrule(lr){4-5}
     & ADE$\downarrow$ & FDE$\downarrow$ & ADE$\downarrow$ & FDE$\downarrow$ \\
    \midrule
    Full (Clustering + Classifier + Experts) & \textbf{0.79} & \textbf{1.09} & \textbf{0.35} & \textbf{0.43} \\
    Random Scene Labels + Classifier        & 1.05 & 2.15 & 0.75 & 1.55 \\
    No Clustering (Direct Scheduling)       & 0.95 & 2.05 & 0.60 & 1.35 \\
    Random Expert Assignment                & 1.05 & 2.30 & 0.80 & 1.65 \\
    Uniform Distribution over Experts       & 0.92 & 2.00 & 0.65 & 1.40 \\
    Single Best Model (SocialMOIF)         & 0.81 & 1.66 & 0.43 & 1.07 \\
    LED + EqMotion + SingularTraj          & 0.82 & 1.68 & 0.44 & 1.10 \\
    Exclude Diffusion Models                & 0.84 & 1.74 & 0.48 & 1.15 \\
    \bottomrule
  \end{tabular}
\end{table}

\paragraph{Hyperparameter Sensitivity Analysis.}
To determine the optimal number of scene categories, we conduct hyperparameter tuning experiments varying $K$ from 3 to 10.
\textbf{Table~\ref{tab:hyperparam}} reports the classification accuracy and trajectory prediction performance across different values of $K$.
Smaller $K$ (e.g., $K=3$) yields high classifier accuracy (91.2\%) but merges distinct scene types, leading to suboptimal prediction performance.
Larger $K$ (e.g., $K=8, 10$) produces noisier clustering with degraded classification accuracy and prediction performance.
Empirically, $K=5$ strikes the best balance between granularity and stability, achieving 84.2\% classification accuracy and the best ADE on both ETH-UCY (0.79) and SDD (0.35).
This confirms that five scene categories provide sufficient expressiveness to capture scene heterogeneity without introducing excessive noise.

\begin{table}[t]
  \centering
  \caption{Hyperparameter sensitivity analysis for the number of scene categories $K$ on ETH-UCY and SDD.
  $K=5$ achieves the best trade-off between classification accuracy and prediction performance.
  \textbf{Bold} indicates the best results.}
  \label{tab:hyperparam}
  \begin{tabular}{cccc}
    \toprule
    \multirow{2}{*}{$K$ (Scene Categories)} &
    \multirow{2}{*}{\shortstack{Classifier\\Accuracy (\%)}} &
    \multicolumn{2}{c}{ADE$\downarrow$} \\
    \cmidrule(lr){3-4}
     & & ETH-UCY & SDD \\
    \midrule
    3  & 91.2 & 0.85 & 0.47 \\
    4  & 87.5 & 0.82 & 0.40 \\
    5  & \textbf{84.2} & \textbf{0.79} & \textbf{0.35} \\
    6  & 81.3 & 0.81 & 0.38 \\
    7  & 78.6 & 0.84 & 0.42 \\
    8  & 76.1 & 0.88 & 0.45 \\
    10 & 72.4 & 0.92 & 0.49 \\
    \bottomrule
  \end{tabular}
\end{table}

\paragraph{Scene Interpretability Analysis.}
\label{sec:case-study}
To qualitatively understand what SceneSelect learns, we analyze the cluster-level statistics of key features in \textbf{Table~\ref{tab:scene}}.
The five discovered scene types exhibit distinct characteristics:
Dense Crowd scenes show extremely high neighbor counts and minimal inter-agent distances;
Medium Speed scenes have moderate density with higher average speeds;
Sparse Static scenes are characterized by very low motion variance and curvature;
Low Density scenes feature sparse agents with high dynamics;
and Complex Motion scenes combine high speeds, large curvature, and dense interactions, presenting the most challenging prediction scenarios.

\begin{table}[t]
  \centering
  \caption{Scene-level feature statistics for the five discovered clusters on ETH-UCY.
  Neighbor Count, Mean Speed, and Trajectory Curvature are defined in \textbf{Equations~\eqref{eq:density}, \eqref{eq:speed}, and \eqref{eq:curvature}}, respectively.
  We highlight the most complex case (\textit{Complex Motion}) with a darker background and the simplest case (\textit{Sparse Static}) with a lighter background.}
  \label{tab:scene}
  \begin{tabular}{lccccccc}
    \toprule
    Scene & \shortstack{Neighbor\\Count} & \shortstack{Mean\\Speed} & \shortstack{Speed\\Std.} & \shortstack{Max\\Speed} & \shortstack{Trajectory\\Curvature} & \shortstack{Close\\Neighbors} & \shortstack{Average\\Distance} \\
    \midrule
    Dense Crowd    & 45.33 & 0.34  & 0.14  & 0.61  & 0.35  & 2.83 & 0.012 \\
    Medium Speed   & 15.84 & 0.45  & 0.12  & 0.68  & 0.35  & 1.16 & 0.0087 \\
    \cellcolor{gray!15}Sparse Static  & \cellcolor{gray!15}18.53 & \cellcolor{gray!15}0.068 & \cellcolor{gray!15}0.037 & \cellcolor{gray!15}0.12 & \cellcolor{gray!15}0.12 & \cellcolor{gray!15}1.76 & \cellcolor{gray!15}0.035 \\
    Low Density    & 11.05 & 0.45  & 0.17  & 0.77  & 0.45  & 1.29 & 4.60 \\
    \cellcolor{gray!30}Complex Motion & \cellcolor{gray!30}32.59 & \cellcolor{gray!30}0.69 & \cellcolor{gray!30}0.44 & \cellcolor{gray!30}1.53 & \cellcolor{gray!30}0.99 & \cellcolor{gray!30}1.50 & \cellcolor{gray!30}0.011 \\
    \bottomrule
  \end{tabular}
\end{table}

We further inspect the confusion matrix of the scene classifier on ETH-UCY.
The diagonal entries are strong (ranging from 80.9\% to 89.1\%), confirming that scenes are reliably recognized, while most confusions occur between conceptually related types (e.g., C1 and C3 both involve high-speed motion).
Across clusters, SceneSelect learns intuitive expert preferences:
more expressive but slower models are favored in complex high-interaction scenes, while lighter-weight experts suffice in sparse or quasi-static scenarios.
This interpretability is valuable for diagnosing system behavior and for communicating model decisions to practitioners.

\begin{figure}[!ht]
  \centering
  \includegraphics[width=\textwidth]{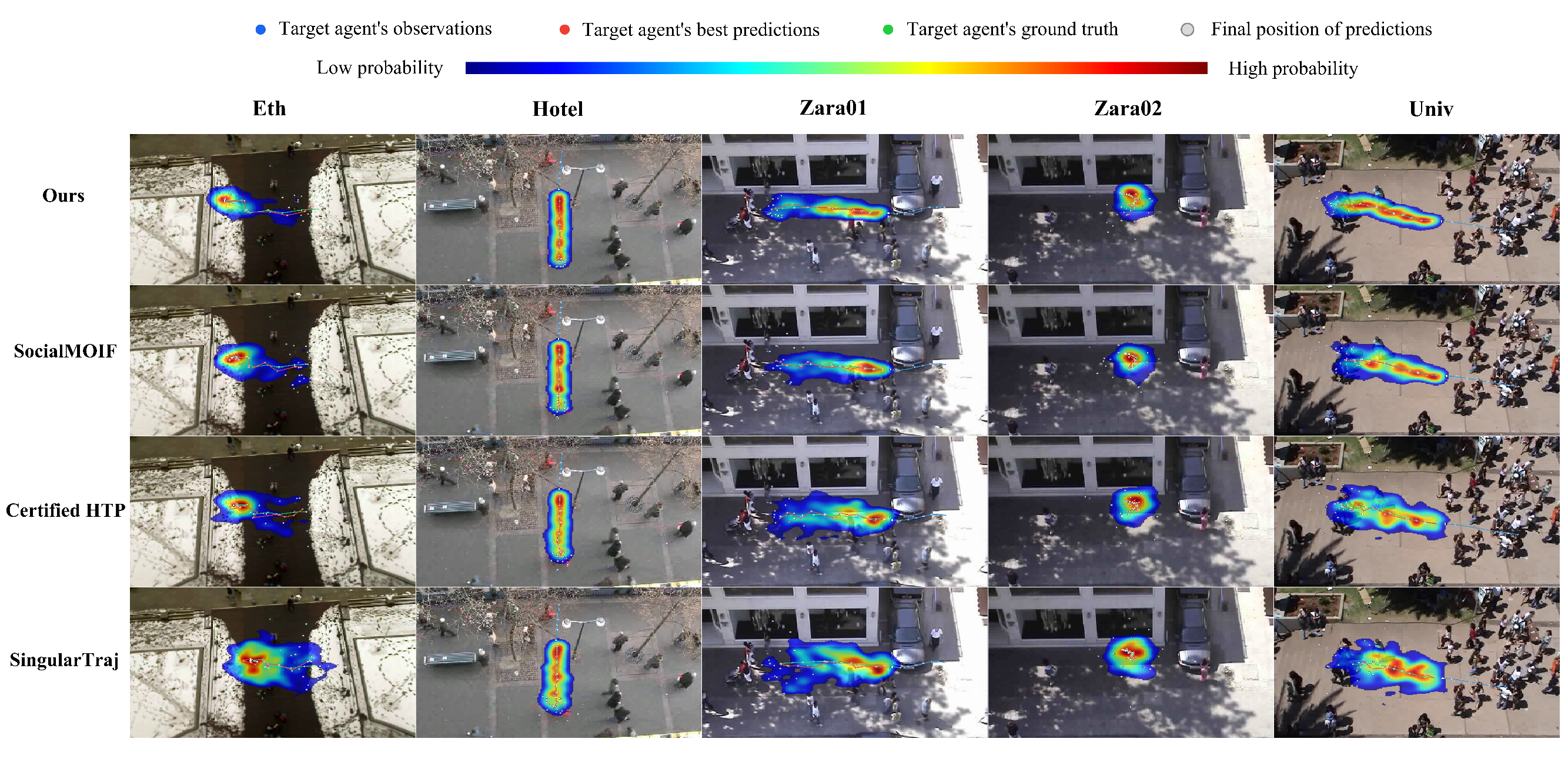}
  \caption{Qualitative trajectory prediction comparison on ETH-UCY dataset across five scenes (Eth, Hotel, Zara01, Zara02, Univ).
  Each row shows predictions from a different method (Ours, SocialMOIF, Certified HTP, SingularTraj).
  Heat maps visualize prediction probability distributions (blue = low, red = high).
  Blue circles indicate observed trajectories, red circles mark best predictions, green circles show ground truth, and gray circles denote final predicted positions.
  SceneSelect (Ours) produces sharper, more focused distributions that better align with ground truth across scene heterogeneity, demonstrating effective scene-aware expert routing.}
  \label{fig:qualitative}
\end{figure}

\paragraph{Qualitative Visualization.}
To better demonstrate the practical benefits of scene-aware expert routing, we conduct qualitative visualization experiments comparing prediction distributions across diverse scenarios.
\textbf{Figure~\ref{fig:qualitative}} presents trajectory predictions on five representative ETH-UCY scenes.
Our method produces sharper, more concentrated probability distributions that align closely with ground truth, particularly in complex scenarios (Hotel, Zara01) where baseline methods generate overly dispersed predictions.
This demonstrates that scene-aware expert routing enables more confident and accurate forecasts by dispatching inputs to specialists optimized for those specific scene characteristics.

\section{Conclusion}

This paper addresses the inherent generalization bottleneck of relying on a single unified model for all scenarios in trajectory prediction by introducing SceneSelect, a pre-trained trajectory scene classification and expert scheduling head utilizing a decoupled Mixture of Experts (MoE) design.
By shifting the research focus from architecture design to scene-level selection learning, SceneSelect discovers an interpretable taxonomy of trajectory scenes, learns to recognize them efficiently, and dispatches real-time inputs to the most suitable expert.
Experiments on three public benchmarks demonstrate that our approach consistently outperforms strong single-model and ensemble baselines, achieving an average improvement of 10.5\% across all datasets.

The main strengths of our approach include: i) decoupling scene understanding from trajectory forecasting enables reusing existing predictors without retraining; ii) the learned scene taxonomy is interpretable and aligned with human-understandable traffic patterns; and iii) the modular design makes it easy to retrofit into existing systems.

Constrained by the strict latency limits of autonomous systems, we deliberately constructed our SceneSelect taxonomy using discrete features rather than exhaustive end-to-end continuous representations.
This reflects a practical trade-off, prioritizing interpretability and immediate plug-and-play deployability over marginal performance gains.
Future work will explore expanding to continuous feature spaces.

We believe that pre-trained selection heads such as SceneSelect offer a promising path toward scalable, modular, and interpretable trajectory prediction systems for future real-world deployments.

\bibliographystyle{splncs04}
\bibliography{references}

\end{document}